\documentclass{ecai} 

\usepackage{latexsym}
\usepackage{amssymb}
\usepackage{amsmath}
\usepackage{amsthm}
\usepackage{booktabs}
\usepackage{color}
\usepackage{tabularray}
\usepackage{multirow}
\usepackage{enumitem}
\usepackage{color}
\usepackage{graphicx}
\usepackage{subcaption}
\usepackage{array} 
\usepackage{float}
\usepackage{hyperref}
\usepackage{afterpage}

\newcommand{\BibTeX}{B\kern-.05em{\sc i\kern-.025em b}\kern-.08em\TeX}

\begin{document}


\begin{frontmatter}


\paperid{123} 


\title{MoSAiC: Multi-Modal Multi-Label Supervision-Aware Contrastive Learning for Remote Sensing}


\author[A]{\fnms{Debashis}~\snm{Gupta}}
\author[A]{\fnms{Aditi}~\snm{Golder}}
\author[B]{\fnms{Rongkhun}~\snm{Zhu}}
\author[C]{\fnms{Kangning}~\snm{Cui}}
\author[C]{\fnms{Wei}~\snm{Tang}}
\author[A]{\fnms{Fan}~\snm{Yang}}
\author[A]{\fnms{Ovidiu}~\snm{Csillik}}
\author[A]{\fnms{Sarra}~\snm{Alaqahtani}}
\author[A]{\fnms{V. Paul}~\snm{Pauca}}

\address[A]{Wake Forest University, NC, USA \\
Email: \{guptd23,golda24,yangfan,csilliko,sarra-alqahtani,paucavp\}@wfu.edu}

\address[B]{Xidian University, China \\
Email: \{zhur\}@wfu.edu}

\address[C]{City University of Hong Kong, Hong Kong \\
Email: cuij@wfu.edu, wetang7-c@my.cityu.edu.hk}

\begin{abstract}
Contrastive learning (CL) has emerged as a powerful paradigm for learning transferable representations without the reliance on large labeled datasets. Its ability to capture intrinsic similarities and differences among data samples has led to state-of-the-art results in computer vision tasks. These strengths make CL particularly well-suited for Earth System Observation (ESO), where diverse satellite modalities such as optical and SAR imagery offer naturally aligned views of the same geospatial regions. However, ESO presents unique challenges, including high inter-class similarity, scene clutter, and ambiguous boundaries, which complicate representation learning—especially in low-label, multi-label settings. Existing CL frameworks often focus on intra-modality self-supervision or lack mechanisms for multi-label alignment and semantic precision across modalities. In this work, we introduce MoSAiC, a unified framework that jointly optimizes intra- and inter-modality contrastive learning with a multi-label supervised contrastive loss. Designed specifically for multi-modal satellite imagery, MoSAiC enables finer semantic disentanglement and more robust representation learning across spectrally similar and spatially complex classes. Experiments on two benchmark datasets, BigEarthNet V2.0 and Sent12MS, show that MoSAiC consistently outperforms both fully supervised and self-supervised baselines in terms of accuracy, cluster coherence, and generalization in low-label and high-class-overlap scenarios.
\end{abstract}

\end{frontmatter}

\section{Introduction}

The contrastive learning (CL) framework has proven particularly revolutionary in the realm of computer vision~\cite{chen2020simple}, offering a powerful avenue for acquiring rich and transferable visual representations without the often prohibitive costs and limitations associated with large-scale labeled datasets. By learning to discern between semantically related and distinct data points, CL enables models to capture intrinsic visual similarities and differences, leading to representations that can be effectively leveraged and fine-tuned for a wide array of downstream tasks. A significant advantage of this self-supervised approach lies in its ability to achieve state-of-the-art performance while substantially reducing the reliance on meticulously annotated data, a bottleneck in many real-world applications.

The inherent characteristics of Earth System Observation (ESO) have made contrastive learning a particularly compelling area of exploration within this community. ESO encompasses the acquisition of vast amounts of data concerning our planet through a diverse suite of satellite-borne and airborne imaging modalities. These include multispectral and hyperspectral optical imagery, synthetic aperture radar (SAR), and light detection and ranging (LiDAR), each providing unique perspectives on the Earth's surface. In this context, a specific geographical location, defined by its longitude, latitude, and a particular point in time, can be naturally considered an anchor. The various available imaging modalities capturing this location at or near that time can then be viewed as distinct, yet inherently related, views of the same underlying Earth surface state. This natural alignment makes ESO data an ideal candidate for contrastive learning strategies. Detection, classification, and segmentation are fundamental machine learning tasks within ESO, essential for the identification and quantitative analysis of critical ecological patterns and the monitoring of temporal changes driven by global warming and human activities~\cite{Schneider}.

However, the application of machine learning to ESO presents unique challenges that distinguish it from the scenarios often studied using benchmark datasets like ImageNet. Several salient characteristics necessitate careful consideration in the design and evaluation of learning algorithms. These include (1) the subtle visual similarity between naturally occurring objects (e.g., distinguishing between different tree species or differentiating deforestation scars from natural landslides), (2) the presence of significant superposition and background clutter (e.g., overlapping canopy in dense forests obscuring ground features), and (3) the inherently fuzzy and ill-defined boundaries of many objects of interest (e.g., the gradual transition between dry and wet sand in deforested areas or the varying growth levels in naturally occurring reforestation). These real-world challenges underscore the need for robust and nuanced machine learning techniques capable of handling the inherent ambiguities and complexities of ESO data.
\begin{figure*}
    \centering
    \includegraphics[width=\linewidth]{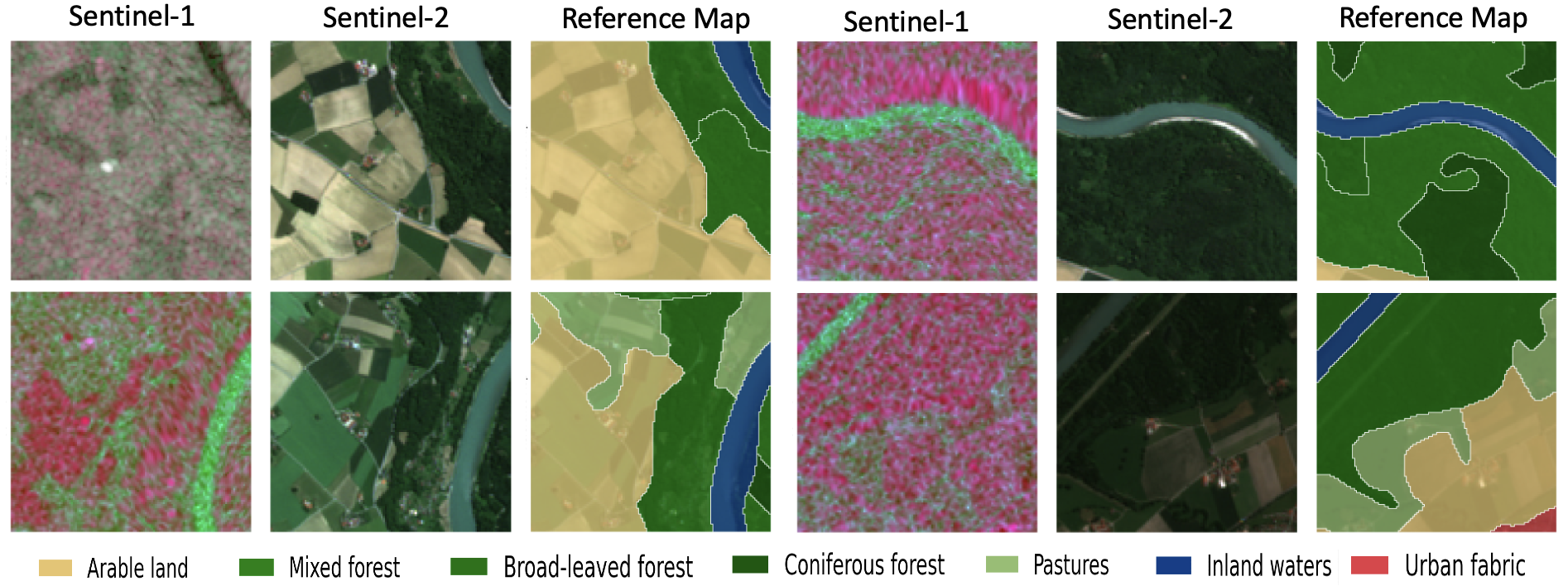}
    \caption{Illustrates the representations of Sentinel-1 and Sentinel-2 along with the Reference Map for the BigEarthnetV2.0 Dataset. The legends are same across the the maps. }
    \label{fig:enter-label}
\end{figure*}

The contrastive learning paradigm was popularized by SimCLR~\cite{chen2020simple}, which learns representations by maximizing agreement between augmented views of the same image within a single modality. While effective for learning invariances, SimCLR is limited to intra-modality settings and does not generalize well to multi-modal data, which is central to Earth System Observation (ESO). To address this, Jain et al.\cite{jain2022multimodal}, Swope et al.\cite{swope2021representation}, and Scheibenreif et al.\cite{scheibenreif2022contrastive} adapted SimCLR for satellite imagery by constructing cross-modality positive pairs from different sensors (e.g., Sentinel-1 and Sentinel-2). However, these methods remain purely self-supervised and struggle with fine-grained semantic distinctions in complex, multi-label settings. Prexl et al.\cite{IaI} extended this direction by combining intra- and inter-modality contrastive objectives and introducing geospatially-aware hard negative mining. While improving alignment in latent space, their method lacks label supervision and offers limited semantic precision in multi-label tasks. More recently, joint optimization of self-supervised and supervised contrastive learning has gained traction for improving generalization and robustness in low-label regimes. A key contribution is MulSupCon by Zhang et al.~\cite{zhang2024multi}, which introduces a label-overlap-aware contrastive loss tailored to multi-label data, enabling finer semantic structuring. However, MulSupCon targets general multi-label classification and does not address remote sensing-specific challenges such as multi-modality, geospatial alignment, or subtle spectral variability. 

In this paper, we specifically investigate the efficacy of joint contrastive self-supervised and fully supervised learning within the challenging context of ESO classification tasks. Our work aims to explore how the benefits of contrastive representation learning can be effectively integrated with supervised learning to improve the accuracy and robustness of segmenting complex environmental features from multi-modal ESO data. Unlike the approach taken by Zhang et al. \cite{zhang2024multi}, which does not target remote sensing or Earth observation data, our method explicitly addresses the complexities inherent in these domains, such as the spatial and temporal characteristics of remote sensing data and the need to handle multiple data modalities. Our contributions are as follows:
\begin{itemize}
\item We propose \textbf{MoSAiC}, a unified framework that jointly optimizes intra- and inter-modality contrastive learning with a multi-label supervised contrastive loss, explicitly designed for complex, multi-modal Earth System Observation (ESO) data.

\item We develop a multi-label, multi-modal loss to geospatial, multi-modal remote sensing tasks, enabling fine-grained semantic alignment across modalities while preserving the discriminative structure within each modality.

\item We conduct extensive experiments on benchmark ESO datasets (BigEarthNet V2.0 (reBEN) and Sent12MS) and demonstrate that MoSAiC achieves superior performance over fully supervised models and existing contrastive baselines in low-label and high-class-overlap regimes, with improved cluster coherence and classification accuracy.
\end{itemize}

\section{Related Work}

Contrastive learning (CL) has emerged as a powerful self-supervised technique for representation learning, especially in domains where annotated data is scarce. SimCLR~\cite{chen2020simple} remains a foundational framework in this space, relying solely on augmented intra-modality views to learn invariant representations. However, such intra-modality objectives are insufficient for multi-modal remote sensing, as they fail to capture cross-sensor semantics essential for joint SAR and optical analysis. Recent work has extended contrastive learning to the remote sensing domain by leveraging the natural alignment of co-registered multi-modal imagery. Jain et al.\cite{jain2022multimodal} adapted the SimCLR framework to Sentinel-1/2 data, forming positive pairs across modalities using spatially aligned patches. Prexl et al.\cite{prexl2023multi} further advanced this direction by introducing a multi-objective formulation that combines both inter-modality and intra-modality contrastive losses, thereby enhancing both sensor-specific robustness and cross-modal alignment. Despite their contributions, these approaches remain unsupervised and lack the semantic grounding necessary to differentiate spectrally similar classes—an inherent challenge in EO datasets like BigEarthNet\cite{kondylatos2025probabilistic}.

To address this, supervised contrastive learning frameworks such as SupCon~\cite{khosla2020supervised} and its multi-label extension MulSupCon~\cite{zhang2024multi} incorporate label supervision to structure the feature space semantically. MulSupCon introduces a label-overlap-aware loss that accounts for partial correlations in multi-label scenarios, significantly improving performance where class boundaries are ambiguous. However, these methods are often confined to unimodal settings or downstream fine-tuning pipelines, limiting their effectiveness in complex sensor fusion tasks.

\textit{Our work} bridges this gap by proposing a hybrid contrastive learning framework that jointly optimizes unsupervised intra-modality invariance and multi-label supervised inter-modality alignment. This design enables both the preservation of modality-specific cues (e.g., SAR backscatter or optical reflectance) and the learning of semantically consistent embeddings across modalities. In contrast to prior work, we explicitly leverage partial label information during contrastive training, rather than deferring it to a downstream classifier. This integrated supervision allows our model to disentangle semantically similar classes—such as Broad-leaved Forest and Coniferous Forest—that are otherwise challenging due to spectral proximity~\cite{fuller2023croma}.

\section{ESO  Data Characteristics and Problem Statement}
Accurate and efficient classification of land cover from Earth System Observation (ESO) data is critical for monitoring environmental change driven by climate dynamics and human activity. This work focuses on the multi-label land-cover classification problem using geo-located, multi-modal satellite imagery. Specifically, we utilize active C-band Synthetic Aperture Radar (SAR) data from Sentinel-1 and passive multispectral optical imagery from Sentinel-2 (bands B2, B3, B4, B8), both at 10-meter resolution. Each image pixel is geo-referenced with latitude and longitude, allowing imagery from different modalities that capture the same location to be treated as distinct but semantically aligned views of the same Earth surface sample.

The classification task is inherently multi-label: each geographic region—represented as a patch of co-registered multi-modal imagery—can be associated with up to 19 land cover types~\cite{sumbul2021bigearthnet}, derived from the 2018 Corine Land Cover map~\cite{feranec2016european}. The goal is to develop models that can accurately predict the set of relevant land cover classes from the input imagery. A key challenge in this domain is the dynamic and evolving nature of land cover, influenced by weather, seasonal shifts, human intervention (e.g., deforestation, mining, urbanization), and other environmental processes~\cite{Schneider}. This temporal variability, combined with the continuous acquisition of vast quantities of ESO data, creates a persistent shortage of labeled examples necessary for fully supervised learning.

To address these challenges, our objective is to investigate the effectiveness of self-supervised learning (SSL), fully supervised learning (SL), and hybrid strategies that combine both. We aim to develop robust, label-efficient models that leverage large-scale unlabeled, geo-located, multi-modal data to improve classification accuracy, generalization, and adaptability in the face of real-world ambiguity, data scarcity, and Earth system variability.

\section{Methodology}

Our methodology integrates self-supervised and fully supervised learning to tackle multi-label land-cover classification from multi-modal Sentinel-1 (S1) and Sentinel-2 (S2) imagery. We design a unified framework that supports both sequential (self-supervised pretraining followed by supervised fine-tuning) and hybrid approaches that jointly optimize contrastive and supervised objectives. At its core, the framework learns robust feature representations through contrastive learning and multi-label supervision, which are used to predict the set of land cover labels for each geographic region.

\subsection{Feature Extraction}
Given $S1$ and $S2$ image patches as input, we employ separate feature encoder networks, $f_{S1}$ and $f_{S2}$, to extract modality-specific feature representations:
\begin{eqnarray}
f_{S1}: S1 \rightarrow h_{S1},\\
f_{S2}: S2 \rightarrow h_{S2},
\end{eqnarray}
where $h_{S1}\in \mathbb{R}^{d_{S1}}$
and 
$h_{S2}\in \mathbb{R}^{d_{S2}}$
represent the feature vectors for Sentinel-1 and Sentinel-2, respectively, and $d_{S1}$ and 
$d_{S2}$ are the dimensions of the feature spaces. The feature encoders, $f_{S1}$ and $f_{S2}$, can be implemented using various deep neural network architectures, such as ResNet variants.

\subsection{Contrastive Learning}

To learn representations without relying on labeled data, we employ a self-supervised contrastive learning strategy. The self-supervised learning objective ($\mathcal{L}_{SSL}$) consists of two intra-modal contrastive loss terms and one inter-modal contrastive loss term:
\begin{equation}
    \mathcal{L}_{SSL} = \mathcal{L}_{intra}(S1) + \mathcal{L}_{intra}(S2) + \mathcal{L}_{inter}(S1, S2),
\end{equation}
\begin{itemize}
    \item 
    \textbf{Intra-modal Contrastive Learning:} The terms $\mathcal{L}_{intra}(S1)$ and $\mathcal{L}_{intra}(S2)$ aim to learn representations that are invariant to different transformations of the input data within each modality. We implement this using the SimCLR framework~\cite{chen2020simple}. 
    For each input image patch from $S1$ and $S2$, we generate two augmented views using a series of data augmentations (e.g., random cropping, random rotation, Gaussian blur). These augmented views are then passed through the corresponding feature encoder ($f_{S1}$ or $f_{S2}$) and a non-linear projection head ($g_{S1}$ or $g_{S2}$) to obtain latent space representations:
    \begin{eqnarray}
        g_{S1}: h_{S1} \rightarrow z_{S1},\\
        g_{S2}: h_{S2} \rightarrow z_{S2},
    \end{eqnarray}
    where $z_{S1} \in \mathbb{R}^{k_{S1}}$ and $z_{S2} \in \mathbb{R}^{k_{S2}}$ are the latent space representations, and $k_{S1}$ and $k_{S2}$ are the dimensions of the latent spaces. 
    The projection heads, $g_{S1}$ and $g_{S2}$, are implemented as multi-layer perceptrons (MLPs) similar to the literature.

    The NT-Xent loss \cite{chen2020simple} is then applied to maximize the similarity between the latent representations of the two augmented views of the same input (positive pair) while minimizing the similarity with the latent representations of augmented views from different inputs (negative pairs) within the same mini-batch.

    \item 
    \textbf{Inter-modal Contrastive Learning:}
    The term $\mathcal{L}_{inter}(S1,S2)$ facilitates the learning of a shared representation space between the two modalities. Unlike intra-modal contrastive learning, which relies on augmented views of the same modality, inter-modal contrastive learning leverages the geo-location information inherent in our dataset. Since Sentinel-1 and Sentinel-2 images are geo-located, we treat co-registered image patches from the two modalities (i.e., patches capturing the same geographical region) as positive pairs. The latent representations $z_{S1}$  and $z_{S2}$
    of these co-registered patches are then encouraged to be similar, while the latent representations of non-corresponding patches from different modalities within the same mini-batch are treated as negative pairs. This process effectively performs implicit data fusion by aligning the representations of different modalities in a shared semantic space.
\end{itemize}

\subsection{Supervised Learning}
To incorporate label information, we employ two supervised learning approaches: a multi-label supervised contrastive loss and a multi-label binary cross-entropy loss.
\begin{itemize}
    \item 
    \textbf{Multi-label Supervised Contrastive Loss:} 
    We use the Multi-Label Supervised Contrastive (MulSupCon) loss, $\mathcal 
    {L}_{msc}$, proposed by~\cite{zhang2024multimodal}, which extends the SupCon loss~\cite{khosla2020supervised} to handle multi-label data. Given a set of latent representations and their corresponding multi-label vector $y$, MulSupCon encourages representations with similar labels to be close to each other in the latent space.

    We apply MulSupCon in two complementary ways:
    \begin{enumerate}
    \item 
    \textbf{Fused Modalities:}
    We concatenate the latent representations of the two modalities $z_{S1}$ and $z_{S2}$ to form a fused representation $[z_{S1},z_{S2}] \in \mathbb{R}^{k_{s1} + k_{s2}}$, and then compute the MulSupCon loss with respect to the ground truth labels $y$:
    \begin{equation}
        \mathcal{L}_{msc}([z_{S1}, z_{S2}], y)
    \end{equation}
    This encourages the model to learn a joint representation that captures the relationships between the two modalities and their relevance to the land cover labels.

    \item 
    \textbf{Intra-modal Augmentations:}
    To further enrich the feature representations, we also apply MulSupCon to augmented views within each modality. For each modality, we generate two augmented views $S1'$ and $S1''$ for Sentinel-1, and $S2'$ and $S2''$ for Sentinel-2. We then compute the MulSupCon loss for each modality's augmented views:
    \begin{eqnarray}
        \mathcal{L}_{msc}([z_{S1'}, z_{S1''}], y) \\
        \mathcal{L}_{msc}([z_{S2'}, z_{S2''}], y)
    \end{eqnarray}
    This encourages the encoders to learn representations that are robust to intra-modal variations while being discriminative with respect to the land cover labels.
    \end{enumerate}

\item 
\textbf{Multi-label Binary Cross-Entropy Loss:}
We also employ a standard multi-label classification approach using the binary cross-entropy (BCE) loss, $\mathcal{L}_{bce}$. 
In this way, we perform data fusion at the feature space level. The feature representations $h_{S1}$ and $h_{S2}$ from the feature encoders are concatenated to form a fused feature vector $[h_{S1}, h_{S2}]\in \mathbb{R}^{d_{S1} + d_{S2}}$. This fused vector is then passed through a linear classification head, $g_c$, to predict a multi-label probability distribution:
\begin{equation}
   g_c: [h_{S1}, h_{S2}] \rightarrow p
\end{equation}
where $p\in [0,1]^n$ is the predicted probability vector for the $n$ land cover labels. The BCE loss is then computed between the predicted probabilities p and the ground truth labels $y$:
\begin{equation}
 \mathcal{L}_{bce}(g_c([h_{S1}, h_{S2}]), y)
\end{equation}
\end{itemize}

\subsection{Joint Training and Ablation}
We formulate multiple training strategies by combining self-supervised and supervised loss terms in different ways. These configurations enable systematic evaluation of our framework and support ablation studies that isolate the contributions of each loss component.
\begin{itemize}
    \item 
    \textbf{Intra-SimCLR: $\mathcal{L}_{intra}(S1) + \mathcal{L}_{intra}(S2)$}.
    This configuration applies SimCLR \cite{chen2020simple} independently to each modality. The learned features $h_{S1}$ and $h_{S2}$ from the pre-trained encoders $f_{S1}$ and $f_{S2}$  are then used to train a downstream multi-label classifier $g_c$  (using BCE loss). This approach mirrors a common practice of self-supervised pre-training followed by supervised fine-tuning.

    \item 
    \textbf{IaI-SimCLR: $\mathcal{L}_{intra}(S1) + \mathcal{L}_{intra}(S2) + \mathcal{L}_{inter}(S1, S2)$}.
    This configuration extends Intra-SimCLR by incorporating the inter-modal contrastive loss \cite{prexl2023multi}, $\mathcal{L}_{inter}(S1, S2)$, during pre-training. This encourages the encoders to learn a shared representation space between the two modalities before the supervised classification stage.

    \item 
    \textbf{MoSAiC-1: $\mathcal{L}_{intra}(S1) + \mathcal{L}_{intra}(S2)+ \mathcal{L}_{msc}([z_{S1}, z_{S2}], y) + \mathcal{L}_{bce}(g_c([h_{S1}, h_{S2}]), y)$}.

This hybrid formulation combines self-supervised contrastive learning (via SimCLR) within each modality, with supervised contrastive learning applied to the fused latent representations across modalities. By also including a standard BCE loss on the fused feature space, MoSAiC-1 encourages the learning of both modality-invariant and semantically meaningful features, benefiting from unlabeled and labeled data simultaneously.

    \item 
    \textbf{MoSAiC-2: $\mathcal{L}_{inter}(S1, S2) + \mathcal{L}_{msc}([z_{S1'}, z_{S1''}], y) + \mathcal{L}_{msc}([z_{S2'}, z_{S2''}], y) + \mathcal{L}_{bce}(g_c([h_{S1}, h_{S2}]), y)$}.  In contrast, MoSAiC-2 replaces self-supervised losses with fully supervised contrastive learning within each modality. The MulSupCon loss is applied independently to augmented views of S1 and S2, promoting robust and label-aware structuring of modality-specific representations. Inter-modality contrast ensures alignment across sensor types, while BCE supervision supports downstream classification. This variant prioritizes label-guided feature learning at both intra- and inter-modal levels.

    
\end{itemize}

In both hybrid approaches, a key distinction from the sequential pre-train and fine-tune strategy is that the parameters of the feature encoders $(f_{S1}, f_{S2})$, the projection heads $(g_{S1}, g_{S2})$, and the supervised classification components (within $\mathcal{L}_{msc}$ and $\mathcal{L}_{bce}$) are optimized \textit{\textbf{jointly}}. 
This end-to-end optimization allows for a synergistic interaction between the self-supervised and supervised learning processes, potentially leading to improved performance and generalization. The specific architectures and training hyperparameters for the feature encoders, projection heads, and classification heads will be detailed in the following section.

\section{Experiments and Results }
\subsection{Experimental Setup}

To evaluate the robustness and stability of our proposed framework, we conduct each experiment over four independent runs. For each run, we randomly sample 10\% of the training data using a stratified strategy that ensures inclusion of all rare classes while maintaining a representative class distribution. We report results as the mean $\pm$ standard deviation across these runs to reflect both average performance and variance. To keep the framework lightweight, we adopt ResNet-34 as the default encoder architecture for $f_{S1}$ and $f_{S2}$, though it can be replaced with any state-of-the-art backbone. To have a fair comparison, we kept the same training approach between all of the baseline models and our architecture. 


\subsection{Datasets}

\noindent \textbf{BigEarthNet (reBEN).} We evaluate our approach on the refined BigEarthNet (reBEN) dataset~\cite{bigearthnet2}, a large-scale multi-modal archive designed for deep learning in remote sensing. It comprises 549{,}488 co-registered image pairs from Sentinel-1 (SAR) and Sentinel-2 (multispectral) satellites, with each pair covering a $1200 \times 1200$ square meter area. Sentinel-2 imagery has been atmospherically corrected using Sen2Cor v2.11, yielding higher-quality reflectance profiles~\cite{bigearthnet2}. Each image pair is annotated with scene-level multi-labels derived from the CORINE Land Cover 2018 map. 
We use the original data split: 237{,}871 images for training, 122{,}342 for validation, and 119{,}825 for testing.

\noindent \textbf{SENT12MS.} We also evaluate our framework on the SENT12MS dataset~\cite{sent12ms}, which includes 75{,}849 training and 8{,}110 testing images with multi-labels. This dataset also provides both single-label and multi-label scene annotations across 10 land cover classes, aligned with the IGBP classification scheme. 


\begin{table*}
\centering
\caption{Shown the performance of MoSAiC-1 and MoSAiC-2 compared with baseline models trained on 10\% data samples for BigEarthNetV2.0 and SENT12MS Datasets. The performance metrics are Average Macro Precision ($AP^M$), Average Macro F1 ($AF1^M$), Average Micro Precision ($AP^\mu$), Average Micro F1 ($AF1^\mu$). Best result is highlighted in \textbf{bold} and second highest result is \underline{underlined}. }
\label{tab:10_percent_training}
\resizebox{\textwidth}{!}{%
\begin{tblr}{
  width = \linewidth,
  colspec = {Q[115]Q[119]Q[100]Q[90]Q[87]Q[96]Q[96]Q[102]Q[106]},
  cells = {c},
  cell{2}{1} = {r=8}{},
  cell{2}{3} = {r=4}{},
  cell{2}{4} = {r=8}{},
  cell{2}{5} = {r=8}{},
  cell{6}{3} = {r=2}{},
  cell{8}{3} = {r=2}{},
  cell{10}{1} = {r=8}{},
  cell{10}{3} = {r=4}{},
  cell{10}{4} = {r=8}{},
  cell{10}{5} = {r=8}{},
  cell{14}{3} = {r=2}{},
  cell{16}{3} = {r=2}{},
  vlines = {black},
  vline{1} = {1-2,10}{black},
  hline{1,18} = {-}{0.08em},
  hline{2} = {1,5}{dotted},
  hline{2} = {2-4,6-9}{},
  hline{3-5,7,9,11-13,15,17} = {2,6-9}{},
  hline{6,14} = {2-3,6-9}{dotted},
  hline{8,16} = {2-3,6-9}{dashed},
  hline{10} = {-}{},
}
\textbf{Dataset} & \textbf{Model}   & \textbf{Architecture} & \textbf{Train Data}     & \textbf{Test Data}        & \textbf{$AP^M$}  & \textbf{$AF1^M$} & \textbf{$AP^\mu$} & \textbf{$AF1^\mu$} \\
BigEarthNet V2.0 & Resnet-34        & Fully Supervised      & {23787 \\(10\%)       } & {119825 \\(100\%)       } & 49.95±4.43          & 37.15±3.53          & 61.88±3.86                          & 53.65±2.96                            \\
                 & Resnet-50        &                       &                         &                           & 55.43±1.99          & 45.17±3.65          & 67.55±1.92                          & 62.36±1.35                            \\
                 & Resnet-101       &                       &                         &                           & 45.49±4.96          & 26.81±7.88          & 54.06±13.22                         & 38.99±3.27                            \\
                 & ConvNext V2 Base &                       &                         &                           & 53.85±1.89          & \underline{50.63±1.34}         & 67.67±1.38                          & \underline{67.50±1.89}                           \\
                 & Intra-SimCLR           & Self-Supervised       &                         &                           & 32.71±2.78          & 19.85±2.05          & 38.59±1.57                          & 42.11±1.34                            \\
                 & IaI-SimCLR              &                       &                         &                           & \textbf{67±2.43}    & {44.17±1.93} & {65±1.27}                    & {65.61±2.77}                   \\
                 & MoSAiC-2        & Our Approach       &                         &                           & \underline{66.0±0.73}           & 43±1.02             & \underline{74±0.57}                             & 65±0.81                               \\
                 & \textbf{MoSAiC-1}    &                       &                         &                           & {64.00±0.22} & \textbf{54.46±0.62} & \textbf{74.04±0.12}                 & \textbf{70.67±0.49}                   \\
Sent12MS         & Resnet-34        & Fully Supervised      & {7584 \\(10\%)       }  & {8110 \\(100\%)       }   & 26.55±0.55          & 25.40±2.37          & 46.52±0.86                          & 58.00±1.59                            \\
                 & Resnet-50        &                       &                         &                           & 29.39±2.91          & 22.40±4.29          & 51.39±5.48                          & 66.72±2.02                            \\
                 & Resnet-101       &                       &                         &                           & 29.43±2.86          & 22.42±4.31          & 51.43±5.44                          & 56.77±2.07                            \\
                 & ConvNext V2 Base &                       &                         &                           & 47.00±3.13          & 44.91±2.76          & 66.76±3.96                          & 67.62±2.08                            \\
                 & Intra-SimCLR           & Self-Supervised       &                         &                           & 21.52±4.34          & 25.33±1.25          & 23.62±1.32                          & 34.68±1.67                            \\
                 & IaI-SimCLR             &                       &                         &                           & 51.64±1.88          & 50.31±1.23          & 72.21±1.34                          & 74.12±1.18                            \\
                 & MoSAiC-2         & Our Approach       &                         &                           & \underline{53.24±0.15}          & \underline{51.40±0.89}          & \underline{73.02±0.43 }                         & \underline{74.51±0.58}                          \\
                 & \textbf{MoSAiC-1}    &                       &                         &                           & \textbf{56.83±0.55} & \textbf{54.27±0.73} & \textbf{74.52±0.58}                 & \textbf{76.26±0.49}                   
\end{tblr}
}
\end{table*}
\subsection{Evaluation Metrics}

In line with established practices in remote sensing~\cite{you2023multi}, we report both macro-averaged and micro-averaged precision and F1-scores. \textbf{Macro-averaging} treats each class equally by computing metrics independently per class and then averaging them. This is crucial for overall system performance, ensuring that rare land cover types—prevalent in long-tail distributions—are not neglected. In contrast, \textbf{micro-averaging} aggregates all true positives, false positives, and false negatives across classes before computing the metrics. This provides a global view of the model’s predictive performance, especially appropriate in scenarios with high class imbalance. 
These metrics align with evaluation protocols used in the BigEarthNet V2.0 benchmark~\cite{bigearthnet2}.

\subsection{Baselines}
We evaluate our framework against two groups of baselines. The first group consists of fully supervised CNN-based models: ResNet-34, ResNet-50, ResNet-101~\cite{resnet}, and ConvNeXt-V2~\cite{convnextv2}. These architectures span a spectrum of complexity, from the lightweight ResNet-34 to the deeper bottleneck-based ResNet-101. ConvNeXt-V2, a modern convolutional architecture, offers strong performance comparable to Swin Transformers while retaining CNN-like efficiency and inductive biases—particularly beneficial in low-label regimes. The second group includes leading contrastive learning approaches: SimCLR~\cite{chen2020simple}, which performs intra-modality self-supervised learning, and IaI~\cite{IaI}, which combines intra- and inter-modality contrastive objectives for multi-modal representation learning. This twofold comparison allows us to assess how MoSAiC performs relative to both traditional supervised pipelines and state-of-the-art contrastive learning frameworks.
\subsection{Results and Analysis}
In realistic EO scenarios, the scarcity of annotated data becomes particularly acute for multi-modal and multi-label tasks, exacerbating the challenge of effectively classifying complex land cover categories. To rigorously evaluate this scenario, we conducted experiments simulating data scarcity by randomly selecting only 10\% of training samples from the two datasets.

\subsubsection{Overall Classification Accuracy.}
Table~\ref{tab:10_percent_training} presents comprehensive results demonstrating that our proposed methods consistently outperform all baselines across both datasets. Specifically, MoSAiC-1 achieves the highest performance across key metrics, including Macro Average Precision, Macro F1-score, Micro Average Precision, and Micro F1-score. Moreover, it shows significantly lower standard deviation, underscoring its robustness and reliability in low-data conditions \cite{zhang2024multimodal}. 
Although IaI achieves marginally higher macro and micro precision (approximately 2\%) on the BigEarthNetV2.0 dataset, it exhibits notably lower F1-scores, highlighting an imbalance between precision and recall. This imbalance is particularly critical for multi-label EO classification tasks, where accurately recognizing diverse land cover types within a single image is essential \cite{patel2021evaluating}. 
Furthermore, on the SENT12MS dataset, both MoSAiC-1 and MoSAiC-2 significantly surpass both SimCLR and IaI models across all evaluation metrics, reinforcing the effectiveness of hybrid supervised-contrastive frameworks in multi-modal contexts. 

\begin{figure}
    \centering
    \includegraphics[width=\linewidth]{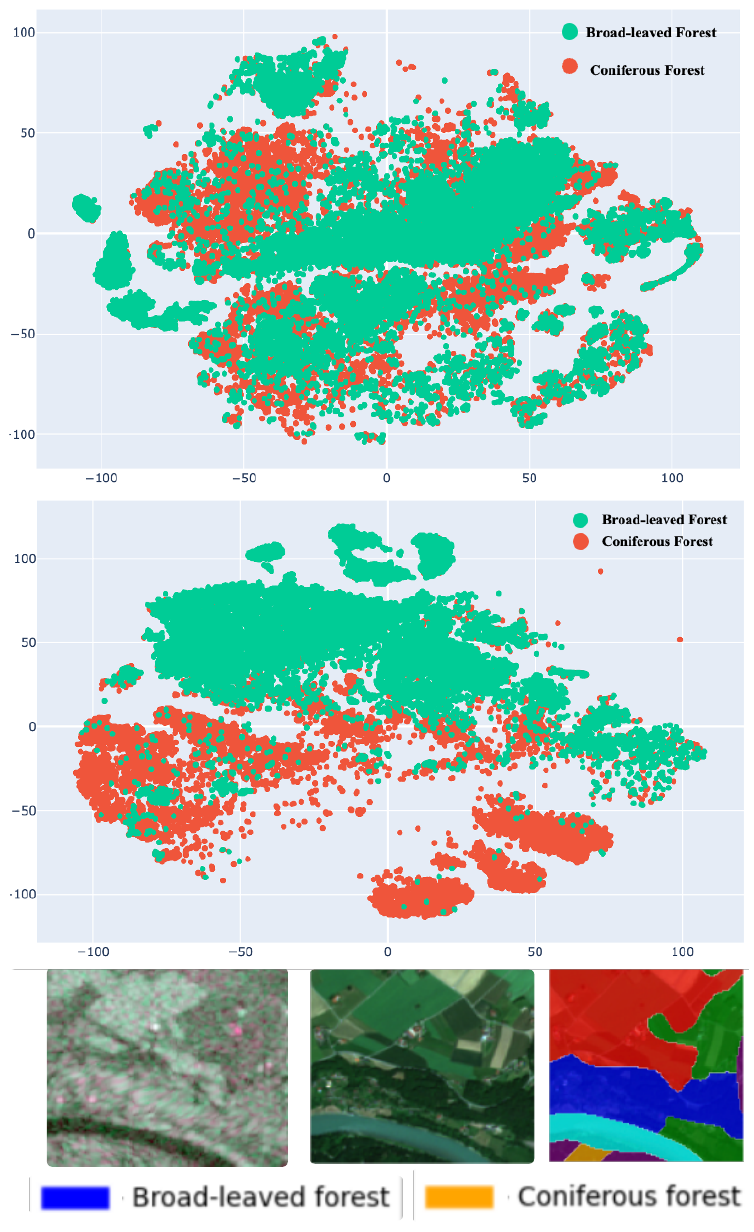}
    \caption{t-SNE projections of multi-modal feature embeddings from Sentinel-1 and Sentinel-2. Top Figure: Without supervised inter-modality loss, class boundaries remain entangled. Middle Figure: With our proposed multi-label supervised contrastive loss, clusters become more structured and semantically aligned, demonstrating improved cross-modal representation learning.}
    \label{fig:feature_projectionl}
\end{figure}
In addition, we performed a t-SNE analysis on the fused feature representations produced by MoSAiC-1 and the IaI-SimCLR baseline. See Figure~\ref{fig:feature_projectionl}. The plots show that MoSAiC-1 achieves better clustering than in the feature space than IaA-SimCLR for two similar classes (Broad-leaved Forest and Coniferous Forest).
Note IaI exhibits large overlap and poorly separated clusters, suggesting limited discriminative structure. These results highlight the advantage of incorporating supervised signals into contrastive learning to guide the formation of more meaningful and robust feature representations.


\begin{table*}[htbp]
\centering
\caption{Hamming Loss (mean ± std) comparison across selected similar and dissimilar classes. The full table is in the Appendix.}
\label{tab:hamming_focused}
\begin{tblr}{
  width = \linewidth,
  colspec = {Q[82]Q[76]Q[76]Q[76]Q[76]Q[76]Q[76]Q[76]},
  column{even} = {c},
  hline{1,9} = {-}{0.08em},
  hline{2,6} = {-}{0.05em},
  font = \small,
}
\textbf{Class}          & \textbf{MoSAiC-1} & \textbf{Intra-SimCLR} & \textbf{IaI-SimCLR} & \textbf{ResNet34} & \textbf{ResNet50} & \textbf{ResNet101} & \textbf{ConvNext} \\
\textbf{Similar}        &                   &                        &              &                   &                   &                    &                   \\
{Broad-leaved \\forest} & 0.19 ± 0.00       & 0.32 ± 0.02            & 0.23 ± 0.02  & 0.30 ± 0.01       & 0.29 ± 0.02       & 0.38 ± 0.10        & 0.34 ± 0.01       \\
{Coniferous \\forest}   & 0.12 ± 0.00       & 0.51 ± 0.02            & 0.14 ± 0.01  & 0.31 ± 0.01       & 0.28 ± 0.04       & 0.32 ± 0.01        & 0.33 ± 0.00       \\
{Mixed \\forest}        & 0.17 ± 0.00       & 0.55 ± 0.03            & 0.22 ± 0.01  & 0.35 ± 0.01       & 0.33 ± 0.03       & 0.35 ± 0.02        & 0.36 ± 0.01       \\
\textbf{Dissimilar}     &                   &                        &              &                   &                   &                    &                   \\
Arable land             & 0.17 ± 0.00       & 0.45 ± 0.04            & 0.22 ± 0.02  & 0.41 ± 0.07       & 0.35 ± 0.07       & 0.37 ± 0.06        & 0.41 ± 0.08       \\
Inland waters           & 0.06 ± 0.00       & 0.62 ± 0.15            & 0.11 ± 0.01  & 0.13 ± 0.01       & 0.13 ± 0.01       & 0.13 ± 0.01        & 0.13 ± 0.00       
\end{tblr}
\end{table*}

\subsubsection{Per-Class Classification Accuracy}
In addition to overall classification, we are also interested in characterizing per-class classification performance. This is particularly relevant for ESO applications that target specific broad classes like vegetation, water, and soil, which often involve finer-grained sub-classes (e.g., broad-leaved, coniferous, and mixed forests). The high spectral similarity among these sub-classes makes accurate differentiation and label assignment challenging.

Hamming loss~\cite{wu2020multi} and the  Brier score~\cite{kruppa2014probability} are two widely used metrics to evaluate performance in multi-label classification tasks. The Brier score is particularly useful for evaluating the accuracy of probabilistic predictions. 
Thus, we compute both {Hamming loss} and {Brier score} for MoSAiC-1, Intra-SimCLR, IaI-SimCLR, ResNet34, ResNet50, ResNet101, and ConvNext. Full results across all 19 classes in the BigEarthNet dataset are provided in the Appendix (due to space limitations). As can be observed, MoSAiC-1 consistently yields the lowest score for both metrics, demonstrating its ability to produce more discriminative and semantically aligned feature representations. 

\begin{figure}
        \centering
        \includegraphics[width=\linewidth]{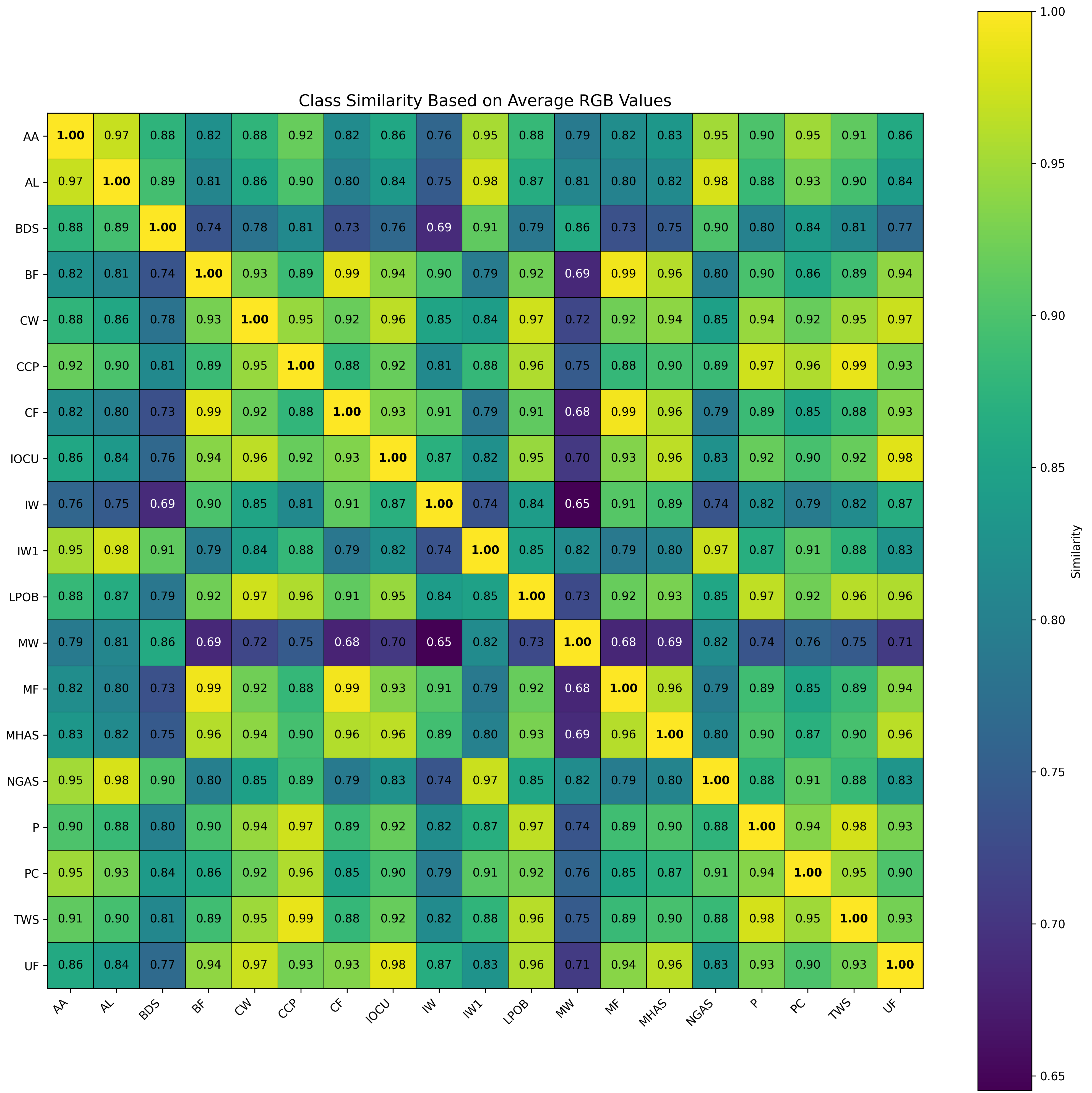}
        \caption{Class similarity matrix based on the Euclidean distance of their spectral bands.}
        \label{fig:class_similarity}
\end{figure}

Furthermore, we analyze class similarity among the 19 BigEarthNet classes using Euclidean distance. Specifically, we compute the Euclidean distance among Sentinel-2 pixels corresponding to each class in the dataset and compute $\mbox{similarity} = 1 / (1 + \mbox{mean Euclidean distance})$, which are reported in Figure~\ref{fig:class_similarity}.
Notice the high similarity scores between most classes. Some class pairs—such as Broad-leaved Forest (BF) and Coniferous Forest (CF), Pastures (PA) and Natural Grassland (NG)—exhibit over 97\% similarity. 
Table~\ref{tab:hamming_focused} shows the Hamming loss calculated for these similar classes as well as for two dissimilar classes. 

\vspace{5px}

\begin{figure}
    \centering
    \includegraphics[width=\linewidth]{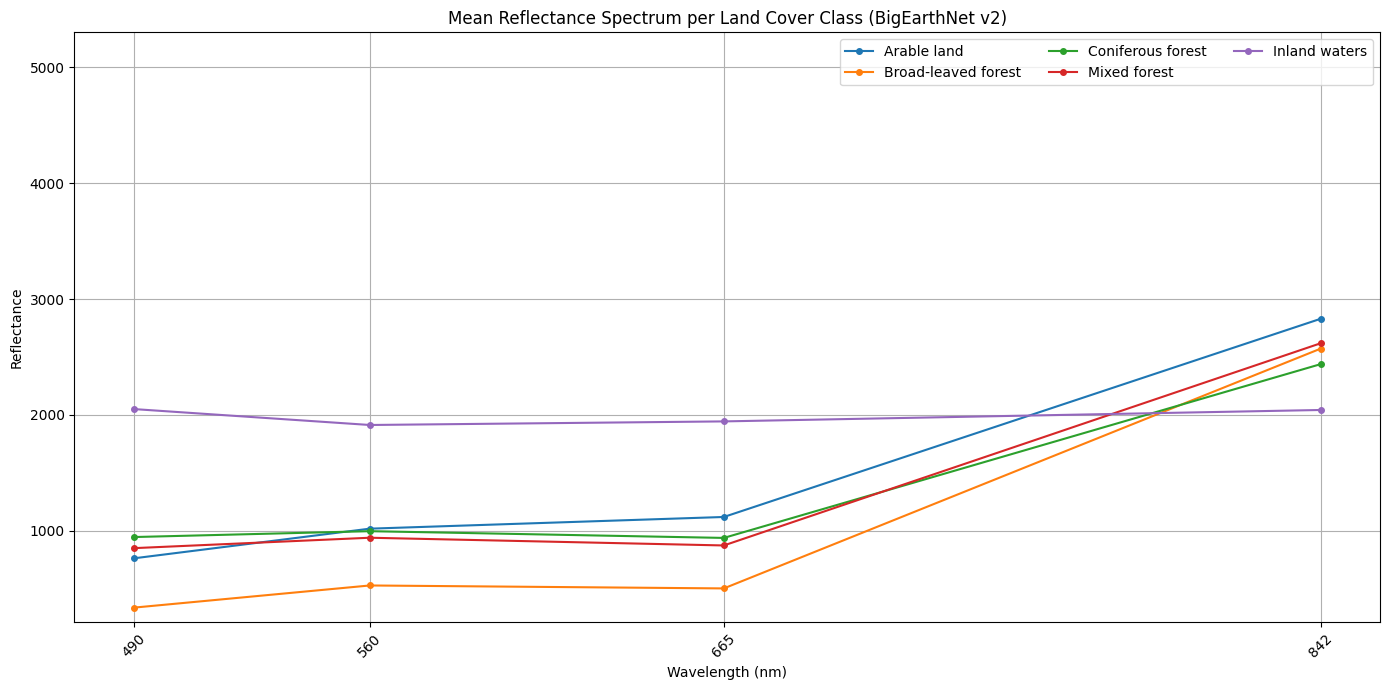}
    \caption{Spectral reflectance traces for pixel clusters of 5 different classes in the BigEarthNet dataset.}
    \label{fig:wavelength}
\end{figure}

The Hamming loss results in the Appendix (and to a certain extent also those in Table~\ref{tab:hamming_focused}) exhibit an interesting performance difference between classes. For some class types the Hamming loss is approximately 0.00, implying that they are more easily identifiable than others whose loss values are relatively higher. Compare, for example, Broad-leaved forest with Coastal wetlands. 
We attribute these differences to the spectral similarities between some classes relative to others. This can be observed in Figure~\ref{fig:wavelength}, which shows the spectral reflectance traces of pixel clusters belonging to 5 classes at bands B2 (490 nm), B3 (560 nm), B4 (665 nm) and B8 (842 nm). 
\vspace{.2in}
Notice the relative difference between the spectral trace for Inland waters versus the other classes.

\section{Limitations and Discussion}

While MoSAiC delivers strong results in multi-modal, multi-label land cover classification, several limitations point to future improvements. First, our spatial alignment strategy assumes that only co-located imagery (e.g., Sentinel-1 and Sentinel-2 of the same region) should be pulled together in the feature space. In reality, similar land-cover scenes may appear in distant locations. Treating such scenes as negatives can harm representation quality. Incorporating similarity-aware contrastive objectives—based on shared land cover distributions or feature-level affinities—could yield smoother, semantically consistent embeddings without requiring supervision. Second, we use simple feature concatenation for fusion and a shallow two-layer projection head. While efficient, this limits the model’s ability to capture richer inter-modality interactions and deeper feature abstractions. Future work may explore advanced fusion techniques such as cross-attention or bilinear pooling, and deeper architectures including transformers, particularly when paired with more annotated data. Despite these limitations, MoSAiC offers a solid foundation for combining supervised and self-supervised contrastive learning in remote sensing. Its strong performance in low-label and high-overlap settings highlights its potential for scalable, label-efficient Earth observation.

\section{Conclusion}
We introduced MoSAiC, a unified framework for multi-label land cover classification from multi-modal Earth observation imagery. By jointly optimizing intra- and inter-modality contrastive learning with supervised multi-label objectives, MoSAiC effectively disentangles subtle semantic differences and enhances feature discriminability. Experiments on BigEarthNet V2.0 and Sent12MS show that our method outperforms fully supervised and contrastive baselines, especially in low-label and high-class-overlap scenarios. t-SNE and per-class analyses confirm MoSAiC’s ability to produce more coherent and well-separated feature clusters. These results underscore the value of integrating supervision into contrastive learning for robust and efficient remote sensing applications.

\bibliography{mybibfile}
\clearpage

\section*{Appendix}
\section{List of the primary Labels} \label{apendix:label}
\begin{table}[htbp]
\centering
\caption{Class labels and their abbreviations used in the dataset.}
\label{tab:class_labels}
\begin{tabular}{@{}lp{0.65\columnwidth}@{}}  
\toprule
\textbf{Abbreviation} & \textbf{Class Label} \\
\midrule
AA            & Agro-forestry areas \\
AL            & Arable land \\
BDS           & Beaches, dunes, sands \\
BF            & Broad-leaved forest \\
CW            & Coastal wetlands \\
CCP           & Complex cultivation patterns \\
CF            & Coniferous forest \\
IOCU          & Industrial or commercial units \\
IW            & Inland waters \\
IW2           & Inland wetlands \\
LPOB     & Land principally occupied by agriculture with significant areas of natural vegetation \\
MW            & Marine waters \\
MF            & Mixed forest \\
MHAS     & Moors, heathland and sclerophyllous vegetation \\
NGAS     & Natural grassland and sparsely vegetated areas \\
P             & Pastures \\
PC            & Permanent crops \\
TWS           & Transitional woodland, shrub \\
UF            & Urban fabric \\
\bottomrule
\end{tabular}
\end{table}

\section{Class Similarity}\label{apendix:sim_SENT12MS}
\begin{figure}[H]
    \centering
    \includegraphics[width=\linewidth]{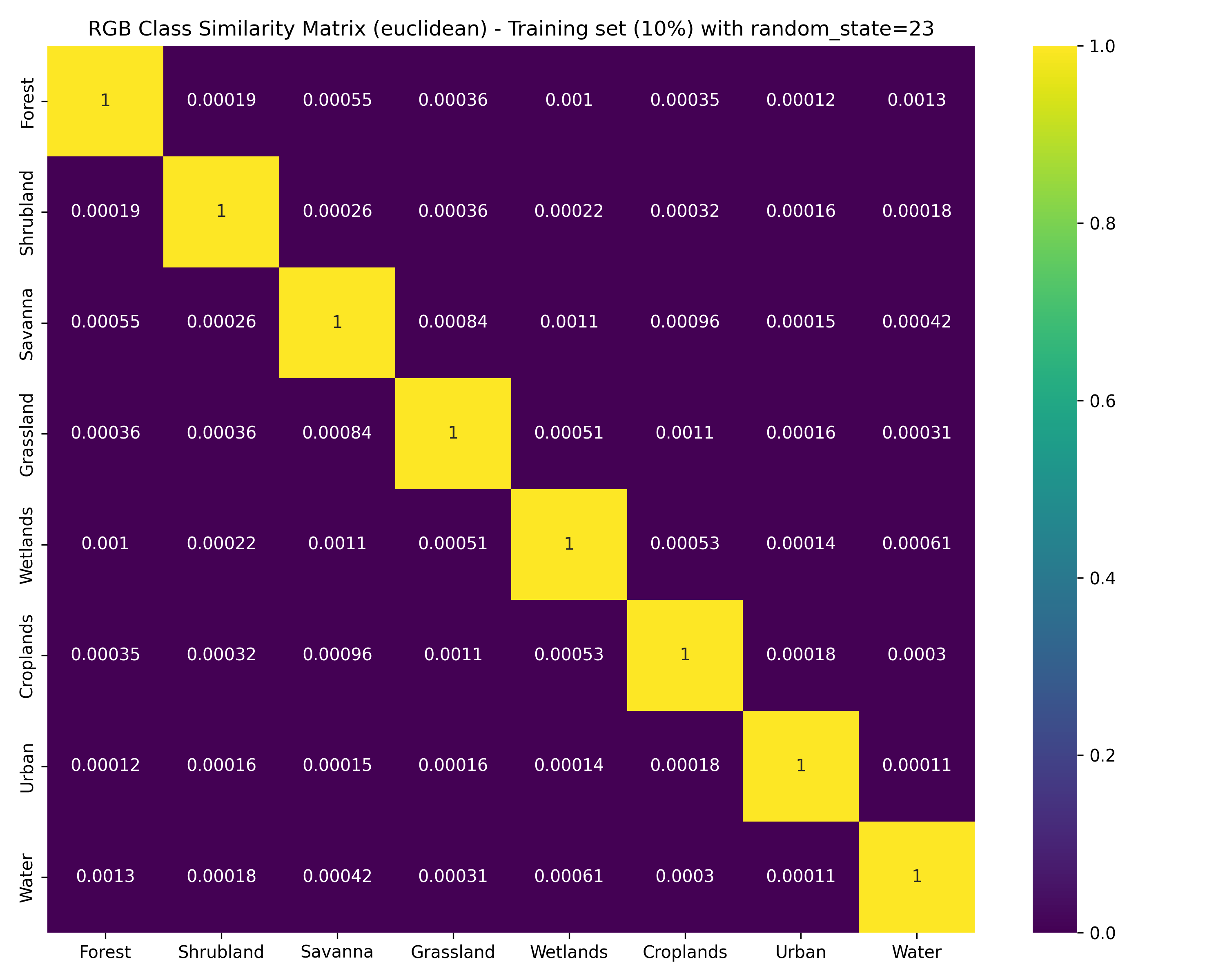}
    \caption{Shown the class similarity based on the calucation of average RGB bands for multi-labels of SENT12MS dataset.}
    \label{fig:enter-label}
\end{figure}

\section{Performance analysis under different modality presence}
To empirically assess the effectiveness and necessity of multimodal data integration for multi-label classification tasks, we conducted a targeted ablation study detailed in Table~\ref{abl:tab_modality}. The primary motivation behind this investigation is to quantitatively demonstrate that combining modalities—Sentinel-1 SAR (S1) and Sentinel-2 multispectral imagery (S2)—yields superior performance compared to relying exclusively on a single modality. Furthermore, we examine simplified multimodal scenarios, where one modality provides detailed features while the other contributes only an average feature representation, simulating situations of limited computational resources or incomplete data availability.
\begin{table}[H]
  \centering
  \caption{Effectiveness of Multimodal Data Integration for Multi-label Classification using MoSAiC for BigEarthNetV2.0 dataset. Here, $AP\rightarrow$Average Precision, $AR\rightarrow$Average Recall, and $AF1\rightarrow$Average F1.}
  \label{abl:tab_modality}
  \resizebox{\linewidth}{!}{%
    \begin{tabular}{|l|c|c|c|c|c|c|}
    \hline
    \textbf{Methodology} & \textbf{$AP^M$} & \textbf{$AR^M$} & \textbf{$AF1^M$} & \textbf{$AP^\mu$} & \textbf{$AR^\mu$} & \textbf{$AF1^\mu$} \\
    \hline
    Only S1 + \textbf{No S2} & 0.14 & 0.12 & 0.07 & 0.26 & 0.18 & 0.16 \\
    Only S2 +\textbf{No S1} & 0.62 & 0.48 & 0.50 & 0.71 & 0.64 & 0.67 \\
    S1+\textbf{Average S2} & 0.34 & 0.14 & 0.13 & 0.53 & 0.21 & 0.30 \\
    S2+\textbf{Average S1} & 0.63 & 0.42 & 0.46 & 0.76 & 0.57 & 0.65 \\
    \hline
    \end{tabular}
  }
\end{table}

\section{Performance on Single Label Classification}
\begin{figure}[H]
    \centering
    \includegraphics[width=\linewidth]{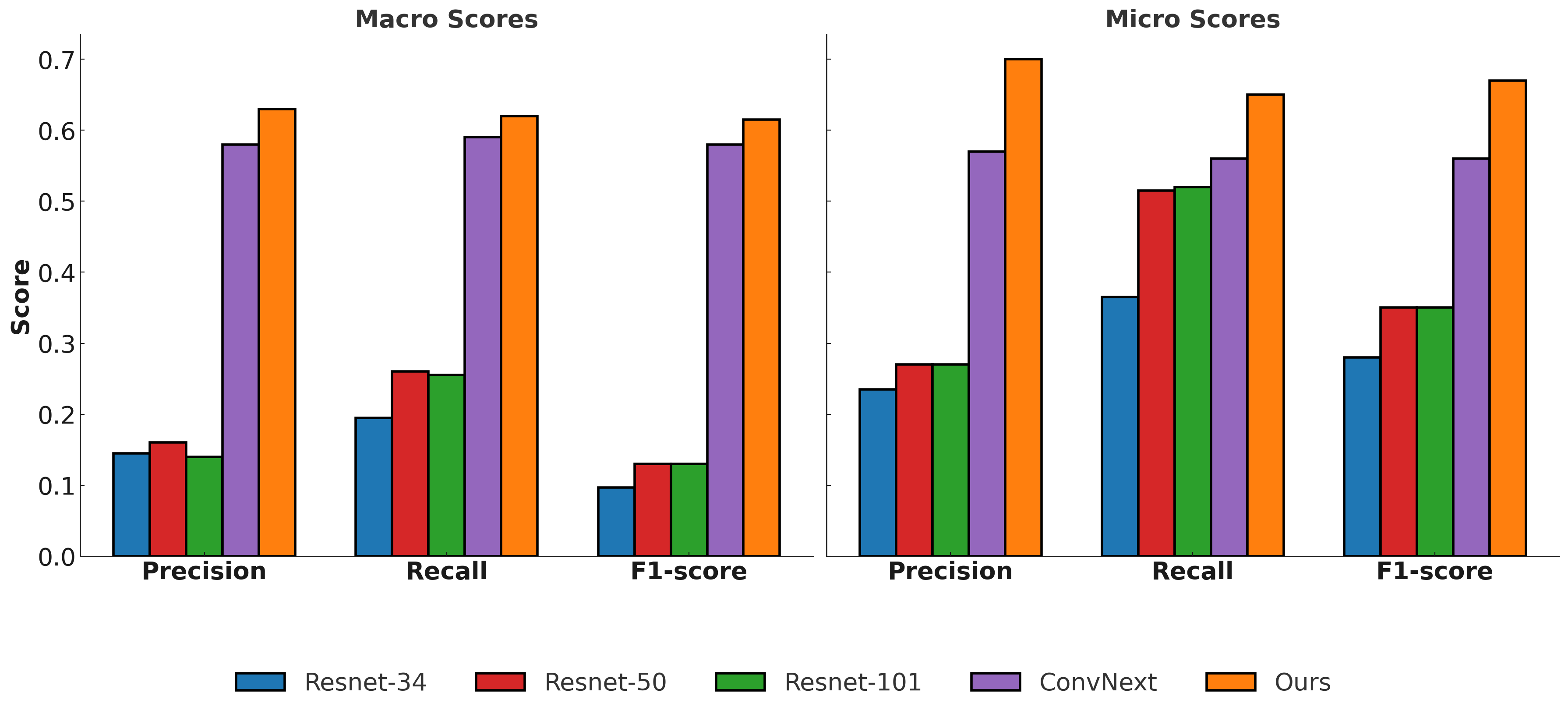}
    \caption{Performance analysis on single label classification.}
    \label{fig:single_label_results}
\end{figure}

The experimental results presented in Table~\ref{fig:single_label_results} clearly illustrate the superior performance of our proposed framework (presented result for MoSAiC-1) over established baseline models, even when evaluated under single-label classification conditions on the SENT12MS dataset. Specifically, our method achieves the highest Macro Precision (0.63), Macro Recall (0.62), Macro F1 Score (0.615), and notably strong Micro Precision (0.70), Micro Recall (0.65), and Micro F1 Score (0.67). This represents a significant improvement over competing models, such as ConvNext and ResNet variants, underscoring the robustness and effectiveness of our approach in extracting discriminative representations from remote sensing imagery. These results validate the versatility of our framework, demonstrating its capability to generalize across both single-label and multi-label scenarios with substantial performance gains.

\section{Hamming Loss}
The Hamming Loss in multi-label classification measures the fraction of labels incorrectly predicted—that is, the average number of mismatches (0 vs. 1 or 1 vs. 0) between the predicted and ground-truth binary label vectors per label per instance.

Let \( \mathbf{y} \in \{0,1\}^L \) be the ground truth vector, and \( \hat{\mathbf{y}} \in \{0,1\}^L \) be the predicted binary labels for an instance (with \( L \) labels). Then for \( N \) samples, the Hamming Loss is:
\begin{equation}
\text{Hamming Loss} = \frac{1}{N \cdot L} \sum_{i=1}^{N} \sum_{j=1}^{L}\left[ y_{ij} \ne \hat{y}_{ij} \right]  
\end{equation}

\begin{table}[H]
\centering
\caption{Per-class Hamming Loss (mean ± std) comparison across different models.}
\label{tab:hamming_distance}
\resizebox{\textwidth}{!}{%
\begin{tabular}{p{7cm}ccccccc}
\toprule
\textbf{Class} & \textbf{MoSAiC-1} & \textbf{Intra-SimCLR} & \textbf{IaI-SimCLR} & \textbf{ResNet34} & \textbf{ResNet50} & \textbf{ResNet101} & \textbf{ConvNext} \\
\midrule
Agro-forestry areas & 0.06 ± 0.01 & 0.07 ± 0.01 & 0.08 ± 0.00 & 0.08 ± 0.00 & 0.09 ± 0.00 & 0.08 ± 0.00 & 0.08 ± 0.00 \\
Arable land & 0.17 ± 0.00 & 0.45 ± 0.04 & 0.22 ± 0.02 & 0.41 ± 0.07 & 0.35 ± 0.07 & 0.37 ± 0.06 & 0.41 ± 0.08 \\
Beaches, dunes, sands & 0.00 ± 0.00 & 0.00 ± 0.00 & 0.00 ± 0.00 & 0.00 ± 0.00 & 0.06 ± 0.05 & 0.00 ± 0.00 & 0.02 ± 0.02 \\
Broad-leaved forest & 0.19 ± 0.00 & 0.32 ± 0.02 & 0.23 ± 0.02 & 0.30 ± 0.01 & 0.29 ± 0.02 & 0.38 ± 0.10 & 0.34 ± 0.01 \\
Coastal wetlands & 0.00 ± 0.00 & 0.00 ± 0.00 & 0.00 ± 0.00 & 0.02 ± 0.02 & 0.00 ± 0.00 & 0.00 ± 0.00 & 0.00 ± 0.00 \\
Complex cultivation patterns & 0.16 ± 0.00 & 0.19 ± 0.01 & 0.17 ± 0.00 & 0.25 ± 0.06 & 0.22 ± 0.02 & 0.19 ± 0.00 & 0.39 ± 0.12 \\
Coniferous forest & 0.12 ± 0.00 & 0.51 ± 0.02 & 0.14 ± 0.01 & 0.31 ± 0.01 & 0.28 ± 0.04 & 0.32 ± 0.01 & 0.33 ± 0.00 \\
Industrial or commercial units & 0.02 ± 0.00 & 0.02 ± 0.00 & 0.02 ± 0.00 & 0.17 ± 0.12 & 0.12 ± 0.01 & 0.05 ± 0.02 & 0.34 ± 0.14 \\
Inland waters & 0.06 ± 0.00 & 0.62 ± 0.15 & 0.11 ± 0.01 & 0.13 ± 0.01 & 0.13 ± 0.01 & 0.13 ± 0.01 & 0.13 ± 0.00 \\
Inland wetlands & 0.04 ± 0.00 & 0.04 ± 0.00 & 0.04 ± 0.00 & 0.04 ± 0.00 & 0.04 ± 0.01 & 0.04 ± 0.00 & 0.04 ± 0.00 \\
Land principally occupied by agriculture with significant areas of natural vegetation & 0.20 ± 0.00 & 0.39 ± 0.04 & 0.21 ± 0.01 & 0.25 ± 0.00 & 0.25 ± 0.00 & 0.26 ± 0.00 & 0.30 ± 0.01 \\
Marine waters & 0.03 ± 0.00 & 0.10 ± 0.00 & 0.04 ± 0.01 & 0.06 ± 0.02 & 0.10 ± 0.06 & 0.08 ± 0.02 & 0.08 ± 0.02 \\
Mixed forest & 0.17 ± 0.00 & 0.55 ± 0.03 & 0.22 ± 0.01 & 0.35 ± 0.01 & 0.33 ± 0.03 & 0.35 ± 0.02 & 0.36 ± 0.01 \\
Moors, heathland and sclerophyllous vegetation & 0.03 ± 0.00 & 0.03 ± 0.00 & 0.03 ± 0.00 & 0.03 ± 0.00 & 0.09 ± 0.04 & 0.03 ± 0.00 & 0.03 ± 0.00 \\
Natural grassland and sparsely vegetated areas & 0.02 ± 0.00 & 0.02 ± 0.00 & 0.02 ± 0.00 & 0.02 ± 0.00 & 0.11 ± 0.02 & 0.05 ± 0.02 & 0.02 ± 0.00 \\
Pastures & 0.13 ± 0.00 & 0.20 ± 0.01 & 0.15 ± 0.01 & 0.29 ± 0.07 & 0.28 ± 0.09 & 0.23 ± 0.00 & 0.24 ± 0.03 \\
Permanent crops & 0.05 ± 0.00 & 0.05 ± 0.00 & 0.05 ± 0.00 & 0.05 ± 0.00 & 0.06 ± 0.01 & 0.05 ± 0.00 & 0.07 ± 0.02 \\
Transitional woodland, shrub & 0.25 ± 0.00 & 0.37 ± 0.02 & 0.28 ± 0.00 & 0.34 ± 0.00 & 0.35 ± 0.00 & 0.37 ± 0.03 & 0.34 ± 0.00 \\
Urban fabric & 0.10 ± 0.01 & 0.10 ± 0.01 & 0.08 ± 0.00 & 0.40 ± 0.26 & 0.28 ± 0.01 & 0.16 ± 0.05 & 0.71 ± 0.03 \\
\bottomrule
\end{tabular}
}
\end{table}

\section{Brier Score}
The Brier Score evaluates the mean squared error between the predicted probabilities and the actual binary labels. It captures both \textit{calibration} (how well predicted probabilities match true likelihoods) and \textit{refinement} (how confident and sharp predictions are).

Given predicted probabilities \( \hat{p}_{ij} \in [0, 1] \) and true binary labels \( y_{ij} \in \{0,1\} \) for each label \( j \) and instance \( i \), the Brier Score is:

\begin{equation}
   \text{Brier Score} = \frac{1}{N \cdot L} \sum_{i=1}^{N} \sum_{j=1}^{L} \left( \hat{p}_{ij} - y_{ij} \right)^2 
\end{equation}

\begin{table}[H]
\centering
\caption{Per-class Brier Score (mean ± std) comparison across different models.}
\label{tab:brier_score}
\resizebox{\textwidth}{!}{%
\begin{tabular}{p{5.5cm}ccccccc}
\toprule
\textbf{Class} & \textbf{MoSAiC-1} & \textbf{Intra-SimCLR} & \textbf{IaI-SimCLR} & \textbf{ResNet34} & \textbf{ResNet50} & \textbf{ResNet101} & \textbf{ConvNext} \\
\midrule
Agro-forestry areas & 0.04 ± 0.00 & 0.06 ± 0.01 & 0.06 ± 0.00 & 0.08 ± 0.00 & 0.08 ± 0.00 & 0.08 ± 0.00 & 0.08 ± 0.00 \\
Arable land & 0.12 ± 0.00 & 0.19 ± 0.01 & 0.18 ± 0.01 & 0.24 ± 0.01 & 0.24 ± 0.02 & 0.27 ± 0.06 & 0.24 ± 0.02 \\
Beaches, dunes, sands & 0.00 ± 0.00 & 0.00 ± 0.00 & 0.00 ± 0.00 & 0.00 ± 0.00 & 0.04 ± 0.03 & 0.00 ± 0.00 & 0.01 ± 0.01 \\
Broad-leaved forest & 0.13 ± 0.00 & 0.26 ± 0.01 & 0.17 ± 0.01 & 0.22 ± 0.01 & 0.23 ± 0.01 & 0.21 ± 0.02 & 0.23 ± 0.02 \\
Coastal wetlands & 0.00 ± 0.00 & 0.00 ± 0.00 & 0.00 ± 0.00 & 0.01 ± 0.01 & 0.00 ± 0.00 & 0.00 ± 0.00 & 0.01 ± 0.00 \\
Complex cultivation patterns & 0.11 ± 0.00 & 0.16 ± 0.01 & 0.13 ± 0.01 & 0.16 ± 0.01 & 0.16 ± 0.00 & 0.15 ± 0.01 & 0.18 ± 0.02 \\
Coniferous forest & 0.09 ± 0.00 & 0.23 ± 0.01 & 0.11 ± 0.01 & 0.24 ± 0.01 & 0.23 ± 0.03 & 0.24 ± 0.04 & 0.25 ± 0.01 \\
Industrial or commercial units & 0.01 ± 0.00 & 0.02 ± 0.00 & 0.02 ± 0.00 & 0.11 ± 0.07 & 0.08 ± 0.02 & 0.03 ± 0.01 & 0.13 ± 0.04 \\
Inland waters & 0.05 ± 0.00 & 0.25 ± 0.11 & 0.09 ± 0.01 & 0.11 ± 0.00 & 0.12 ± 0.01 & 0.12 ± 0.00 & 0.12 ± 0.00 \\
Inland wetlands & 0.03 ± 0.00 & 0.04 ± 0.00 & 0.03 ± 0.00 & 0.04 ± 0.00 & 0.04 ± 0.00 & 0.04 ± 0.00 & 0.04 ± 0.00 \\
Land principally occupied by agriculture, with significant areas of natural vegetation & 0.13 ± 0.00 & 0.17 ± 0.00 & 0.16 ± 0.01 & 0.19 ± 0.00 & 0.21 ± 0.02 & 0.19 ± 0.01 & 0.19 ± 0.00 \\
Marine waters & 0.02 ± 0.00 & 0.10 ± 0.00 & 0.03 ± 0.00 & 0.05 ± 0.02 & 0.07 ± 0.04 & 0.07 ± 0.03 & 0.07 ± 0.02 \\
Mixed forest & 0.12 ± 0.00 & 0.23 ± 0.01 & 0.16 ± 0.01 & 0.26 ± 0.03 & 0.28 ± 0.03 & 0.26 ± 0.05 & 0.25 ± 0.02 \\
Moors, heathland and sclerophyllous vegetation & 0.02 ± 0.00 & 0.03 ± 0.00 & 0.03 ± 0.00 & 0.03 ± 0.00 & 0.06 ± 0.02 & 0.03 ± 0.00 & 0.03 ± 0.00 \\
Natural grassland and sparsely vegetated areas & 0.02 ± 0.00 & 0.02 ± 0.00 & 0.02 ± 0.00 & 0.02 ± 0.00 & 0.07 ± 0.01 & 0.04 ± 0.01 & 0.02 ± 0.00 \\
Pastures & 0.09 ± 0.00 & 0.18 ± 0.00 & 0.12 ± 0.01 & 0.19 ± 0.03 & 0.20 ± 0.04 & 0.18 ± 0.01 & 0.17 ± 0.00 \\
Permanent crops & 0.04 ± 0.00 & 0.05 ± 0.00 & 0.04 ± 0.00 & 0.06 ± 0.01 & 0.05 ± 0.00 & 0.05 ± 0.00 & 0.05 ± 0.00 \\
Transitional woodland, shrub & 0.18 ± 0.00 & 0.21 ± 0.03 & 0.20 ± 0.00 & 0.26 ± 0.03 & 0.26 ± 0.02 & 0.25 ± 0.03 & 0.27 ± 0.00 \\
Urban fabric & 0.07 ± 0.00 & 0.08 ± 0.02 & 0.07 ± 0.00 & 0.33 ± 0.20 & 0.24 ± 0.00 & 0.13 ± 0.02 & 0.47 ± 0.07 \\
\bottomrule
\end{tabular}
}
\end{table}

\section{Feature Heatmap}
We use GradCam to generate the heatmap of the features extracted by our approach, \texttt{MoSAiC}.
\begin{figure}[H]
    \centering
    \includegraphics[width=\linewidth]{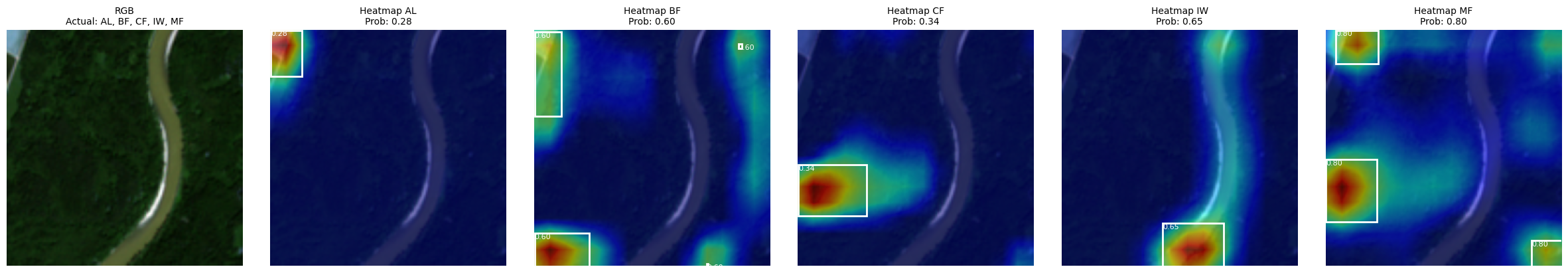}
    \caption{Illustrates the feature heatmap using GradCam for MoSAiC for Sentinel-2 image.}
    \label{fig:enter-label}
\end{figure}

\end{document}